\documentclass{article} % For LaTeX2e
\usepackage{preprint_style,times}

\usepackage{amsmath,amsfonts,bm}

\def\eqref#1{equation~\ref{#1}}
\def\1{\bm{1}}

\DeclareMathAlphabet{\mathsfit}{\encodingdefault}{\sfdefault}{m}{sl}
\SetMathAlphabet{\mathsfit}{bold}{\encodingdefault}{\sfdefault}{bx}{n}

\usepackage{hyperref}
\usepackage{url}

\usepackage{booktabs}
\usepackage{graphicx}
\usepackage{multirow}
\usepackage{booktabs}
\usepackage{multirow}
\usepackage{array}
\usepackage{tabularx}
\usepackage{caption}
\usepackage{wrapfig}

\title{Nexus-VLA: Bridging Human and Robot Experiences via a Shared Hierarchical Semantic Latent Space}

\author{%
\makebox[\dimexpr\textwidth-2\tabcolsep\relax][c]{%
\begin{tabular}{c}
Biaoliang Guan$^{1,*}$,
Tingyu Yuan$^{2,3,*}$,
Wen Ye$^{2,3,*}$,
Ziming Ji$^{4,*}$ \\
Xu Yang$^{2}$,
Yan Huang$^{2,3}$,
Jihua Zhu$^{1,\dagger}$,
Chaoyang Zhao$^{2,\dagger}$ \\[4pt]
$^{1}$XJTU
\qquad
$^{2}$CASIA
\qquad
$^{3}$UCAS
\qquad
$^{4}$BUPT
\end{tabular}%
}%
}

\begin{document}

\maketitle

\begingroup
\renewcommand{\thefootnote}{}
\footnotetext{
$^{*}$ Equal contribution.
\qquad
$^{\dagger}$ Corresponding authors.
}
\endgroup

\fancyhead{}

\begin{abstract}

Large-scale egocentric human videos provide rich and scalable manipulation experience for robot policy learning, yet substantial embodiment gaps in visual appearance, kinematic structure, and action spaces hinder effective knowledge transfer. Existing methods often decouple cross-embodiment knowledge alignment from robot policy learning, making it difficult for human manipulation knowledge to directly and fully contribute to robot action learning. To address this issue, we propose Nexus-VLA, a hybrid policy learning framework that enables cross-embodiment knowledge transfer through end-to-end human-robot joint training, allowing human manipulation experience to directly contribute to robot policy learning. We first introduce a hierarchical cross-embodiment semantic alignment paradigm that progressively establishes shared manipulation semantics from task intent to fine-grained interactions, enabling more complete use of the manipulation knowledge contained in human videos. Building on this hierarchy, we design a shared manipulation latent representation that jointly encodes multi-level shared knowledge into an internal representation directly used for continuous action generation, improving policy learning efficiency while effectively leveraging human knowledge. Across RoboCasa-GR1, OOD settings, and real-world robot experiments, Nexus-VLA achieves an 80.5\% task success rate on RoboCasa-GR1 and demonstrates strong advantages in human-data utilization, learning efficiency, and cross-scenario generalization.

\end{abstract}

\section{Introduction}

Because collecting robot demonstrations at scale remains costly and challenging~\citep{khazatsky2024droid}, massive egocentric human videos, which capture rich real-world manipulation and interaction knowledge, provide a scalable source of data for robot policy learning~\citep{grauman2022ego4d}. However, substantial differences between humans and robots in visual appearance, kinematic structure, and action spaces~\citep{chen2025vidbot} make it challenging to extract manipulation representations that are semantically consistent across embodiments, thereby hindering the effective transfer of human manipulation knowledge to robot policies. Despite these differences, humans and robots share similar manipulation semantics and interaction structures when performing the same task. Therefore, a central challenge in learning robot policies from human videos is to identify such shared representations and transform them into a unified representation that effectively supports robot action generation.

To better exploit human data, recent studies have begun to use embodiment-invariant manipulation semantics in human demonstrations as bridges between humans and robots~\citep{bharadhwaj2023zero}. However, existing approaches often establish human-robot correspondences through only one aspect of semantic consistency, such as aligning high-level task semantics~\citep{kim2026ego} or matching human hand motion with robot end effectors~\citep{chen2026translation}, leaving shared cross-embodiment knowledge only partially modeled. Manipulation semantics are not a single flat signal, but form a hierarchy that progressively connects abstract intent to concrete physical interaction. Manipulation intent describes what to do, spatial semantics specify what to interact with and where, motion semantics characterize how to perform the operation, and interaction semantics further capture how the action affects the object. We argue that these coarse-to-fine semantic levels jointly define the shared structure of human and robot manipulation, suggesting that cross-embodiment correspondence should be progressively established across different semantic levels.

\begin{figure}[t]
    \centering
    \includegraphics[width=1\linewidth]{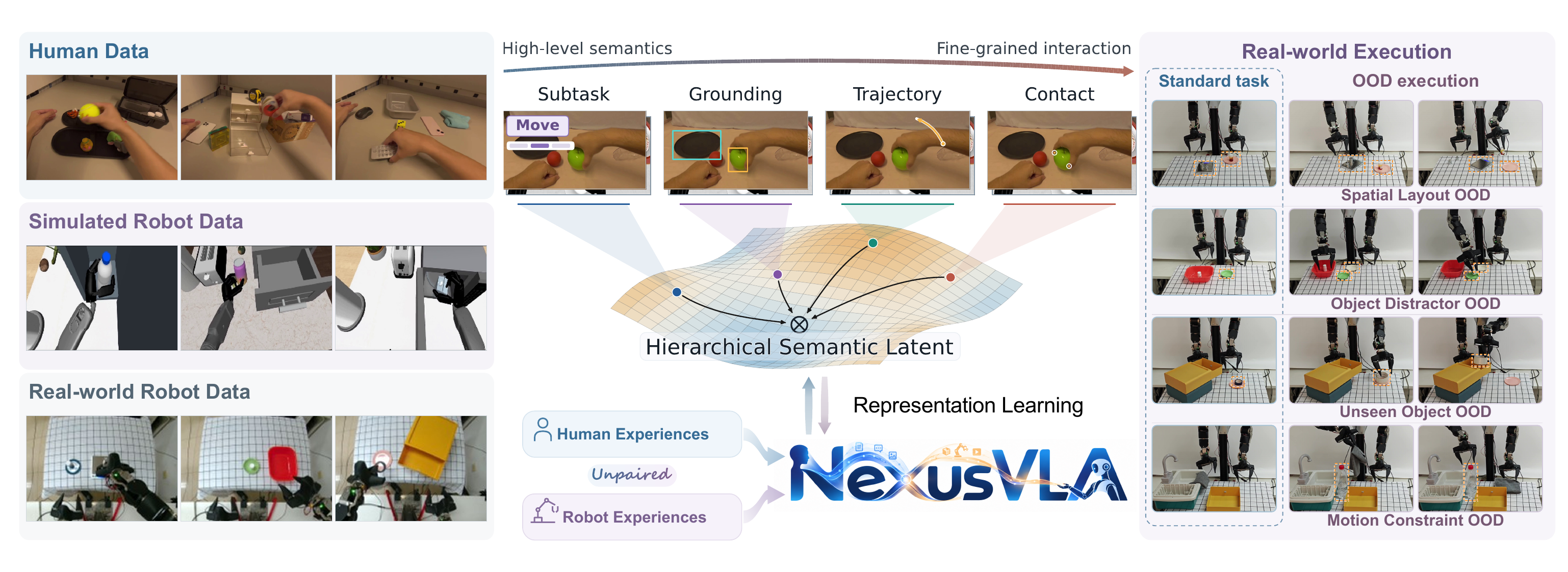}
    \caption{Nexus-VLA bridges human and robot experiences through a Hierarchical Semantic Latent, enabling cross-embodiment knowledge transfer and real-world generalization.}
    \label{fig:intro}
\end{figure}

However, explicitly modeling manipulation semantics separately at different levels is insufficient to capture their underlying structure. These semantics are not independent; rather, they describe the same manipulation process at different levels of abstraction and form a progressively refined, hierarchically constrained structure. Modeling each level in isolation makes it difficult to preserve cross-level dependencies and consistency, motivating a unified representation of multi-level manipulation semantics. However, directly using explicit semantics as intermediate policy representations introduces additional computation and decision latency during inference~\citep{bai2026latent}, while their structured or discrete form is also ill-suited for directly driving continuous action generation~\citep{wei2026libra}. The key challenge is therefore to transform hierarchical manipulation knowledge shared across embodiments into an internal representation suitable for action generation, allowing knowledge from human demonstrations to directly participate in robot policy decision-making. In this way, multi-level knowledge from human demonstrations can be more completely transferred to robot policies, enabling more effective use of human data.

As illustrated in Figure~\ref{fig:intro}, we propose Nexus-VLA, an efficient mixed policy-learning framework for jointly learning from human and robot demonstrations. Nexus-VLA encodes hierarchical manipulation knowledge shared across embodiments into a Hierarchical Semantic Latent that directly participates in action prediction, allowing shared knowledge from human demonstrations to effectively contribute to robot policy learning while improving cross-scenario generalization. Specifically, we first introduce a hierarchical cross-embodiment semantic alignment paradigm. With structured manipulation experts, we generate consistent supervision for human and robot demonstrations across manipulation intent, spatial semantics, motion semantics, and interaction semantics, progressively establishing coarse-to-fine semantic correspondences between the two embodiments. Building on this hierarchy, we aggregate these interdependent shared semantics through joint multi-level prediction objectives into a Semantic Latent, which is combined with vision-language features for action prediction. The robot action learning objective further shapes this latent representation through end-to-end optimization, encouraging it to preserve cross-embodiment semantic consistency while remaining directly useful for action generation. Nexus-VLA jointly trains on mixed human and robot demonstrations in a single stage and, at inference time, directly predicts actions from the Semantic Latent and vision-language features without explicitly generating intermediate semantics.

We systematically evaluate Nexus-VLA on RoboCasa-GR1, multiple out-of-distribution (OOD) scenarios, and real-robot platforms. Nexus-VLA achieves an 80.5\% task success rate on RoboCasa-GR1 and demonstrates strong generalization across OOD tasks and real-robot settings. These results show that connecting human experience with robot action learning through hierarchical semantic latent representations enables more effective use of large-scale human demonstrations and improves both robot policy learning and cross-scenario generalization.

In summary, the main contributions of this paper are as follows:

1. \textbf{An Efficient Single-Stage Co-Training Framework:} We overcome the limitations of traditional methods that rely on paired cross-embodiment data or multi-stage pre-training by realizing end-to-end joint optimization over unaligned human videos and robot demonstrations, which significantly improves both human data utilization and policy learning efficiency.

2. \textbf{A Unified Hierarchical Semantic Latent Representation:} We are the first to encode multi-level manipulation semantics from abstract intents to fine-grained physical interactions into a compact latent variable. This representation retains rich operational structures while bypassing the need for explicit intermediate semantic inference, directly interfacing with continuous action generation.

3. \textbf{Extensive Evaluation Across Simulated and Real-World Environments:} Nexus-VLA achieves an 80.5\% task success rate on the RoboCasa-GR1 benchmark and demonstrates outstanding cross-scene generalization and data efficiency across various out-of-distribution (OOD) scenarios and real-robot platforms.

\section{Related Work}
\label{sec:related_work}

\textbf{Learning Robot Policies from Human Videos.}
Leveraging rich manipulation experience from large-scale egocentric human videos to learn robot policies has been widely explored recently. 
One line of work first pre-trains visual, kinematic, or action-related representations on human videos, and subsequently adapts or fine-tunes the policy using robot demonstrations~\citep{yang2025egovla,ma2026humanscale}. 
Further efforts introduce explicit cross-embodiment alignment stages to facilitate knowledge transfer from humans to robots~\citep{zheng2026egoscale,yu2026dm0}. 
However, such multi-stage pipelines allow human knowledge to act on final policy learning only indirectly. 
Consequently, recent studies have begun to explore direct co-training on mixed human and robot demonstrations~\citep{kareer2025egomimic}, yet straightforward data mixing still struggles to establish consistent cross-embodiment manipulation supervision.

\textbf{Cross-Embodiment Manipulation Alignment.}
To bridge embodiment discrepancies, constructing cross-embodiment stable manipulation representations has emerged as a core strategy for connecting human experience with robot control. 
Existing work typically establishes human-robot correspondences at different levels of manipulation semantics, ranging from high-level task intents~\citep{kim2026ego,lee2026act} and mid-level spatial affordances or geometric grounding~\citep{yang2025magma,xu2025a_}, to low-level end-effector trajectories and physical interactions~\citep{chen2026translation}. 
Other methods narrow the embodiment gap by leveraging paired videos or unified pose spaces~\citep{xie2026human2robot,kareer2025egomimic}. 
However, these approaches generally focus on single-level or local semantic alignment, isolating the inter-level dependencies across semantic granularities and thus limiting the depth and completeness of cross-embodiment knowledge transfer.

\textbf{Latent Manipulation Representations for Robot Control.}
Beyond extracting shared semantics, transforming these semantics into internal representations suitable for robot control represents another core challenge. 
One category of methods explicitly expresses the semantics as structured intermediate information (e.g., affordance maps or predicted trajectories)~\citep{li2025coa,xu2025a_,yang2025magma}. 
While interpretable, such methods require generating explicit intermediate representations during inference, which introduces computational overhead and decision latency; structured signals cannot directly serve as an effective interface for continuous action generation. 
Another line of work encodes manipulation information into compact latent representations~\citep{ye2025latent,chen2025moto,bu2025univla}. 
Although well-suited for end-to-end control, these latent spaces lack explicit multi-level semantic supervision, making it difficult to form structured semantic constraints and weakening their capability to support precise action generation.

\begin{figure}[t]
    \centering
    \includegraphics[width=1\linewidth]{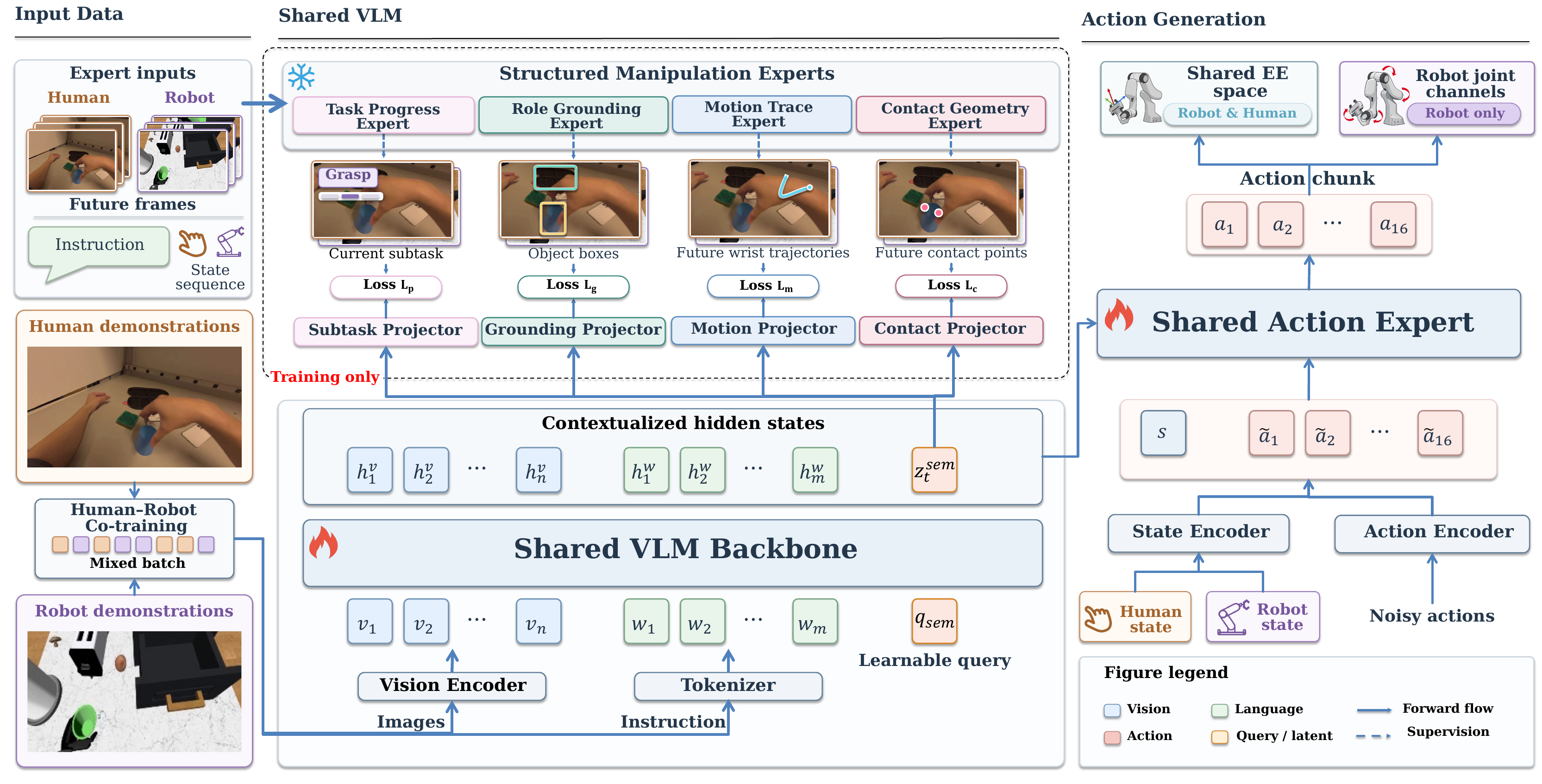}
    \caption{Overview of the Nexus-VLA framework. Hierarchical semantic latents enable joint human–robot learning and action generation.}
    \label{fig:framework}
\end{figure}

\section{Method}
\label{sec:method}

As illustrated in Figure~\ref{fig:framework}, Nexus-VLA performs end-to-end mixed training over unpaired human and robot demonstrations within a unified policy learning framework, enabling human manipulation knowledge to directly contribute to robot action learning. A shared vision-language backbone encodes both types of demonstrations and aggregates four types of cross-embodiment manipulation semantics (Intent, Spatial, Motion, and Interaction) spanning high-level task understanding to low-level physical interaction into a Hierarchical Semantic Latent, forming a unified internal representation for manipulation. We further organize human and robot actions into a unified action representation space, allowing semantic and action supervision to jointly shape the Semantic Latent, which, together with vision-language features, conditions continuous robot action generation.

\subsection{Hierarchically Shaped Shared Manipulation Latent Representation}
\label{sec:hierarchical_latent}

To transform cross-embodiment manipulation knowledge into an internal representation that can directly support action generation, Nexus-VLA constructs a Hierarchical Semantic Latent and shapes it with multi-level manipulation semantics ranging from high-level task understanding to concrete physical interaction.

\subsubsection{Hierarchical Shared Manipulation Semantics}
\label{sec:hierarchical_manipulation_semantics}

Despite differences in visual appearance, kinematic structure, and execution form, humans and robots share similar task goals, object spatial relations, motion processes, and physical interactions when performing the same manipulation. We organize such cross-embodiment shared information into four semantic levels (Intent, Spatial, Motion, and Interaction) according to their level of abstraction, progressively grounding manipulation semantics from high-level task understanding to concrete actions and physical interactions.

Specifically, Intent represents the local task goal, i.e., ``what to do''; Spatial represents the spatial locations of the manipulated object and target region, i.e., ``what to manipulate and where''; Motion represents the motion trend of the acting embodiment toward subsequent interaction, i.e., ``how to move''; and Interaction represents effective interaction locations and contact states, i.e., ``how to interact.'' These four types of semantics can be defined consistently for both human and robot demonstrations, providing shared semantic supervision for cross-embodiment representation learning and subsequently shaping the Hierarchical Semantic Latent.

\subsubsection{Hierarchical Semantic Latent Construction}
\label{sec:semantic_latent}

To provide a unified internal representation for manipulation semantics at different levels, we append a learnable semantic token
$\mathbf q_{\mathrm{sem}}\in\mathbb R^d$
to the end of the vision-language token sequence and define its final hidden state as the Hierarchical Semantic Latent.

Given the visual observation $\mathbf o_t$ at time $t$ and the task instruction $\ell$, let $\mathbf V_t$ and $\mathbf T_t$ denote the corresponding visual and language tokens, respectively. We feed the semantic token together with both inputs into the shared vision-language backbone $F_\theta$, yielding
\begin{equation}
\mathbf H_t
=
\left[
\widetilde{\mathbf V}_t;
\widetilde{\mathbf T}_t;
\mathbf z_t^{\mathrm{sem}}
\right]
=
F_\theta
\left(
\left[
\mathbf V_t;
\mathbf T_t;
\mathbf q_{\mathrm{sem}}
\right]
\right),
\label{eq:semantic_latent}
\end{equation}
where $\widetilde{\mathbf V}_t$ and $\widetilde{\mathbf T}_t$ denote the contextually encoded visual and language features, respectively, and $\mathbf H_t$ denotes the complete multimodal hidden sequence. The final hidden state corresponding to the semantic token,
$\mathbf z_t^{\mathrm{sem}}\in\mathbb R^d$,
is defined as the Semantic Latent.

Because $\mathbf q_{\mathrm{sem}}$ is placed at the end of the multimodal sequence, its hidden state attends to the preceding visual and language context through causal self-attention, forming a compact internal representation of the current manipulation scene and task instruction. Subsequent multi-level semantic prediction objectives jointly shape $\mathbf z_t^{\mathrm{sem}}$, enabling it to further encode cross-embodiment manipulation semantics ranging from high-level task understanding to concrete physical interaction.

\subsubsection{Multi-level Semantic Shaping}
\label{sec:semantic_shaping}

To enable the Semantic Latent to capture manipulation knowledge at different levels of abstraction, we jointly shape it using four semantic targets: Intent, Spatial, Motion, and Interaction. Let
$\mathcal K=\{\mathrm{intent},\mathrm{spatial},\mathrm{motion},\mathrm{interaction}\}$
denote the set of semantic levels. For each semantic level $k\in\mathcal K$, we predict the corresponding semantic target from the same Semantic Latent
$\mathbf z_t^{\mathrm{sem}}$:
\begin{equation}
\hat{\mathbf y}_t^{k}
=
g_k\!\left(\mathbf z_t^{\mathrm{sem}}\right),
\qquad
k\in\mathcal K,
\label{eq:semantic_heads}
\end{equation}
where $g_k$ denotes the projector for the $k$-th semantic level. To ensure that the multi-level supervision primarily shapes the shared latent representation rather than relying on additional modeling capacity from task-specific predictors, we keep each projector lightweight, using it only to map the Semantic Latent to the corresponding supervision space.

Corresponding to the four shared semantics defined in Section~\ref{sec:hierarchical_manipulation_semantics}, the projectors respectively output a local task description, the 2D locations of the manipulated object and target region, future $H$-step bimanual wrist trajectories, and future contact locations and states. All four supervision targets are automatically constructed from demonstrations by offline Structured Manipulation Experts. These experts are used only to generate training labels and are not involved in policy inference; implementation details are provided in the appendix. 

For each semantic level $k$, let $\bar{\ell}_t^k$ denote the prediction loss computed and normalized over valid target elements when necessary. The overall multi-level semantic shaping objective is
\begin{equation}
\mathcal L_{\mathrm{sem}}
=
\sum_{k\in\mathcal K}
\lambda_k\,\bar{\ell}_t^k,
\label{eq:semantic_loss}
\end{equation}
where $\lambda_k$ controls the relative weight of each semantic level. Since all four supervision signals jointly shape the same
$\mathbf z_t^{\mathrm{sem}}$,
the resulting latent captures manipulation semantics across different levels of abstraction. At inference time, these semantic targets need not be explicitly generated; the Semantic Latent is directly used for action generation.

\subsubsection{Semantics-Enhanced Action Generation}
\label{sec:latent_action_generation}

After multi-level semantic shaping, the action generation module uses action tokens to cross-attend to the multimodal hidden sequence $\mathbf H_t$, which contains the Semantic Latent, thereby jointly leveraging visual, language, and multi-level manipulation semantics. Given the noisy action sequence $\mathbf A_t^\tau$, flow timestep $\tau$, and robot proprioceptive state $\mathbf s_t$, the action generation module $G_\phi$ predicts the conditional velocity field:
\begin{equation}
\hat{\mathbf v}_t^\tau
=
G_\phi
\left(
\mathbf A_t^\tau,
\tau,
\mathbf s_t
\,;\,
\mathbf H_t
\right).
\label{eq:semantic_conditioned_action}
\end{equation}
Thus, Intent, Spatial, Motion, and Interaction need not be explicitly generated as intermediate semantics during inference. The complete continuous action formulation and Flow Matching training objective are presented in Section~\ref{sec:joint_training}.

\subsection{End-to-End Human-Robot Mixed Training}
\label{sec:joint_training}

To enable human demonstrations to directly contribute to robot action learning, Nexus-VLA organizes human and robot demonstrations into a unified training action representation and jointly trains them end-to-end within the same policy optimization process.

\subsubsection{Dual-Space Action Representation}
\label{sec:dual_space_action}

Although human demonstrations do not provide robot joint control, human wrists and robot end effectors share a common description of motion relative to the environment. We therefore decompose the training action representation into a cross-embodiment shared wrist/end-effector motion space and a robot-specific joint control space. At time $t$, the unified action representation is defined as
\begin{equation}
\mathbf a_t
=
[\mathbf a_t^{\mathrm{joint}};\mathbf a_t^{\mathrm{wrist}}]
\in
\mathbb R^{d_j+d_w},
\label{eq:dual_space_action}
\end{equation}
where
$\mathbf a_t^{\mathrm{joint}}\in\mathbb R^{d_j}$
denotes the robot joint action, and
$\mathbf a_t^{\mathrm{wrist}}\in\mathbb R^{d_w}$
denotes the bimanual wrist or dual-end-effector motion represented under a unified coordinate system and action parameterization.

For embodiment type
$e\in\{\mathrm{human},\mathrm{robot}\}$,
we define the corresponding action validity masks as
\begin{equation}
\mathbf M^{(\mathrm{human})}
=
[\mathbf 0_{d_j};\mathbf 1_{d_w}],
\qquad
\mathbf M^{(\mathrm{robot})}
=
[\mathbf 1_{d_j};\mathbf 1_{d_w}].
\label{eq:action_mask}
\end{equation}

Accordingly, human demonstrations provide action supervision through the shared wrist/end-effector space, while robot demonstrations additionally provide joint-control supervision, allowing both data sources to jointly train the same action generation model.

During mixed training, we explicitly control the overall sampling ratio between human and robot demonstrations, while sampling datasets within each embodiment source proportionally to their sizes. This decouples the human-robot training ratio from the relative sizes of the underlying datasets.

\subsubsection{Joint Optimization Objective}
\label{sec:joint_optimization}

Nexus-VLA jointly optimizes multi-level semantic shaping and continuous action generation, allowing the Semantic Latent to encode cross-embodiment manipulation semantics while directly supporting robot control.
For continuous action generation, we adopt conditional flow matching. Following the standard formulation, we interpolate a ground-truth action chunk $\mathbf A_t$ with Gaussian noise to obtain $\mathbf A_t^\tau$ at flow timestep $\tau$, with target velocity $\mathbf v_t^\star$. Conditioned on the embodiment state $\mathbf s_t^{(e)}$ and multimodal context $\mathbf H_t$, the action generator predicts the corresponding velocity field $\hat{\mathbf v}_t^\tau$.

To accommodate the different action supervision available for human and robot demonstrations, we extend the embodiment-specific action mask in
~\eqref{eq:action_mask} over the action chunk to obtain
$\mathbf\Pi^{(e)}$ and compute the flow matching loss only over valid action dimensions:
\begin{equation}
\mathcal L_{\mathrm{FM}}
=
\mathbb E
\left[
\frac{
\left\|
\mathbf\Pi^{(e)}
\odot
\left(
\hat{\mathbf v}_t^\tau-\mathbf v_t^\star
\right)
\right\|_2^2
}{
\left\|\mathbf\Pi^{(e)}\right\|_1+\epsilon
}
\right].
\label{eq:masked_flow}
\end{equation}

Finally, the action learning objective is jointly optimized with the semantic shaping objective introduced in Section~\ref{sec:semantic_shaping}:
\begin{equation}
\mathcal L_{\mathrm{joint}}
=
\mathcal L_{\mathrm{FM}}
+
\lambda_{\mathrm{sem}}
\mathcal L_{\mathrm{sem}}.
\label{eq:joint_objective}
\end{equation}
Semantic and action supervision are thus jointly optimized under a single end-to-end objective. At inference time, the model generates action chunks through the learned velocity field and executes the robot joint-action component.

\begin{table*}[t]
\centering
\caption{
Comparison on the RoboCasa-GR1 benchmark.
We report task success rates (\%), averaged within each of
the five task groups. Avg. denotes the unweighted mean across
all 24 tasks. Nexus-VLA is evaluated using the checkpoint
at 150K training steps.
}
\label{tab:robocasa_main}

\small
\setlength{\tabcolsep}{4pt}
\renewcommand{\arraystretch}{1.10}

\begin{tabularx}{\textwidth}{
    @{}
    >{\centering\arraybackslash}m{1.6cm}
    l
    *{6}{>{\centering\arraybackslash}X}
    @{}
}
\toprule
Type & Method
& Close
& Cuttingboard
& Placemat
& Tray
& Plate
& Avg. \\
\midrule

\multirow{2}{*}{\shortstack[c]{Foundation\\VLA}}
& GR00T-N1.6
& 24.2 & 56.9 & 51.9 & 55.1 & 57.6 & 47.6 \\

& ABot-M0
& 61.7 & 61.6 & 51.0 & 56.8 & 58.0 & 58.3 \\

\midrule

\multirow{3}{*}{\shortstack[c]{Semantic\\Enhanced}}
& VP-VLA
& 54.3 & 60.8 & 54.5 & 46.0 & 53.5 & 53.8 \\

& IntentVLA
& 59.7 & 54.8 & 55.5 & 49.6 & 66.5 & 57.0 \\

& ZR-0
& 41.7 & \underline{83.6} & \underline{74.8} & 72.0 & \textbf{84.2} & 69.3 \\

\midrule

\multirow{3}{*}{\shortstack[c]{Human\\Experience}}
& DyPES-VLA
& 55.3 & 63.2 & 51.5 & 62.4 & 64.0 & 59.3 \\

& JoyAI-RA 0.1
& \textbf{71.7} & 68.4 & 45.0 & 68.8 & 55.0 & 63.2 \\

& ACE-Ego-0
& 59.0 & 83.2 & 70.5 & \underline{78.0} & 76.5 & \underline{72.8} \\

\midrule

Ours
& \textbf{Nexus-VLA}
& \underline{70.3}
& \textbf{89.6}
& \textbf{80.0}
& \textbf{83.6}
& \underline{81.0}
& \textbf{80.5} \\

\bottomrule
\end{tabularx}
\end{table*}

\begin{table}[t]
\centering
\caption{
OOD generalization on RoboCasa-GR1 under illumination changes, visual noise, novel spatial configurations and object-receptacle compositions, objects seen only in EgoDex, and fully unseen objects, with 500 rollouts per setting.
}
\label{tab:ood_generalization}
\resizebox{\linewidth}{!}{
\begin{tabular}{lccccccc}
\toprule
Method
& Illumination
& Visual Noise
& Cross-Config.
& Obj.-Receptacle
& Ego-Object
& Fully Unseen
& Avg. \\
\midrule

Baseline
& 62.0
& 58.4
& 51.2
& 45.0
& 40.8
& 31.6
& 48.2 \\

\textbf{Nexus-VLA}
& \textbf{76.8}
& \textbf{73.6}
& \textbf{71.0}
& \textbf{66.8}
& \textbf{69.4}
& \textbf{55.6}
& \textbf{68.9} \\

\bottomrule
\end{tabular}
}
\end{table}

\section{Experiments}

\subsection{Experimental Setup}
\label{sec:experimental_setup}

We use RoboCasa-GR1~\citep{bjorck2025gr00t} as our primary simulation benchmark. Built on RoboCasa~\citep{nasiriany2024robocasa}, the benchmark contains 24 humanoid manipulation tasks using the Fourier GR-1 robot, covering object transport and interactions with articulated objects such as drawers, cabinet doors, and microwaves. In addition to the standard test scenarios, we construct six controlled OOD settings to evaluate generalization under illumination changes, visual noise, spatial configuration changes, novel object-receptacle compositions, objects observed in human demonstrations, and fully unseen objects.

We train the model using 24,000 robot trajectories from RoboCasa-GR1, covering all 24 tasks, together with 21,446 filtered egocentric pick-and-place videos from EgoDex~\citep{hoque2026egodex}. The two data sources are unpaired and jointly used to train a single policy following the human-robot mixed training strategy described in Section~\ref{sec:joint_training}. We train the model for 200K steps with a global batch size of 512 and a learning rate of $3\times10^{-5}$. Training is conducted on two compute nodes, each equipped with eight NVIDIA H20E 141GB GPUs. The number of evaluation rollouts for each experiment is reported alongside the corresponding results.

On RoboCasa-GR1, we compare Nexus-VLA against foundation VLAs (GR00T-N1.6 and ABot-M0~\citep{yang2026abot}), semantic-enhanced methods (VP-VLA~\citep{wang2026vp}, IntentVLA~\citep{lian2026intentvla}, and ZR-0~\citep{li2026training}), and approaches leveraging human experience (DyPES-VLA~\citep{li2026dypes}, JoyAI-RA 0.1~\citep{zhang2026joyai}, and ACE-Ego-0~\citep{li2026ace}). In real-world experiments, we additionally compare with $\pi_{0.5}$~\citep{intelligence2025pi_} and GR00T-N1.6, together with our Baseline and intermediate model variants.

\subsection{Overall Performance and Generalization}
\label{sec:overall_performance}

As shown in Table~\ref{tab:robocasa_main}, on the standard RoboCasa-GR1 test distribution, Nexus-VLA achieves an average success rate of 80.5\% across all 24 tasks, outperforming the strongest baseline, ACE-Ego-0, at 72.8\% by 7.7 percentage points. Across the five task groups, Nexus-VLA achieves the highest success rates on Cuttingboard, Placemat, and Tray, and ranks second on both Close and Plate. These results demonstrate stable in-distribution performance across diverse manipulation tasks.

We further construct controlled OOD tests in RoboCasa-GR1 to evaluate generalization under visual changes, scene composition shifts, and novel objects. As shown in Table~\ref{tab:ood_generalization}, Nexus-VLA achieves an average success rate of 68.9\% across the six OOD settings, compared with 48.2\% for the Baseline, corresponding to a 20.7 percentage-point gap. Under illumination changes and visual noise, Nexus-VLA achieves 76.8\% and 73.6\%, exceeding the Baseline by 14.8 and 15.2 percentage points, respectively. Under novel spatial configurations and object-receptacle compositions, the performance gaps further increase to 19.8 and 21.8 percentage points. These results indicate that the manipulation knowledge learned by Nexus-VLA transfers more robustly to new visual conditions and scene compositions, consistent with its hierarchical semantic modeling of task intent, object roles, and spatial relations.

Object-level generalization further highlights the value of transferring human manipulation experience. For Ego-Objects that are absent from robot training data but observed in human demonstrations, Nexus-VLA achieves a success rate of 69.4\%, exceeding the Baseline by 28.6 percentage points, the largest performance gap among the six OOD settings. Meanwhile, on Fully Unseen objects that appear in neither the human nor robot training data, Nexus-VLA still achieves 55.6\%, exceeding the Baseline by 24.0 percentage points. The former demonstrates that human video experience can effectively complement robot-side unseen objects, while the latter suggests that the hierarchical semantic representation can further support generalization across object instances.

\begin{table}[t]
\centering
\caption{
Robot-only variants are evaluated at 120K training steps, whereas variants using human demonstrations are evaluated at 180K steps with a human-to-robot sampling ratio of 1:2, matching the expected robot-data exposure across variants.
}
\label{tab:ablation}

\small
\setlength{\tabcolsep}{5pt}
\renewcommand{\arraystretch}{1.10}

\begin{tabular}{lcccc}
\toprule
Method
& Human-Robot
& Hierarchical
& Semantic
& Success Rate (\%) $\uparrow$ \\
& Co-training
& Alignment
& Representation
& \\
\midrule

Baseline
& -- & -- & -- & 72.42\%\\

+ Hierarchical Alignment
& -- & \checkmark & -- & 75.67\%\\

+ Human--Robot Co-training
& \checkmark & -- & -- & 71.75\%\\

+ Independent Latents
& -- & \checkmark & Independent & 76.92\%\\

+ Unified Latent
& -- & \checkmark & Unified & 78.25\%\\

\midrule

Nexus-VLA
& \checkmark
& \checkmark
& Unified
& 80.17\%\\

\bottomrule
\end{tabular}
\end{table}

\subsection{Ablation Study and Learning Efficiency}
\label{sec:ablation_efficiency}

As shown in Table~\ref{tab:ablation}, hierarchical semantic modeling consistently improves policy learning. Compared with the Baseline, Hierarchical Alignment increases the success rate from 72.42\% to 75.67\%, a gain of 3.25 percentage points, indicating that multi-level supervision from task intent to fine-grained interaction provides useful structure for manipulation learning. Under the same hierarchical semantic supervision, replacing independent semantic representations with a Unified Semantic Latent further improves the success rate from 76.92\% to 78.25\%. This suggests that the gains arise not only from additional semantic supervision, but also from jointly representing manipulation knowledge across different semantic levels in a form that better supports action generation.

The co-training results further reveal how human data contributes to policy learning. With matched robot-data exposure, direct Human-Robot Co-training converges more slowly (71.75\%), but reaches 76.32\% at 200K steps, surpassing the Robot-only Baseline peak of 72.92\%. In contrast, when Hierarchical Alignment and the Unified Semantic Latent are kept fixed, adding Human-Robot Co-training increases the success rate from 78.25\% to 80.17\%, providing an additional gain of 1.92 percentage points. This controlled comparison suggests that hierarchical semantic alignment and the unified latent representation better organize cross-embodiment shared knowledge, allowing human manipulation experience to contribute more effectively to robot policy learning.

\begin{wrapfigure}{r}{0.48\linewidth}
    \centering
    \includegraphics[width=\linewidth]{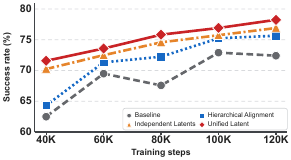}
    \caption{Training efficiency of different semantic modeling strategies.}
    \label{fig:training_efficiency}
\end{wrapfigure}

The training curves further reveal how different semantic modeling strategies affect learning efficiency. As shown in Figure~\ref{fig:training_efficiency}, Hierarchical Alignment consistently outperforms the Baseline throughout training, indicating that hierarchical semantic supervision from task intent to fine-grained interaction provides more effective learning signals for action learning. Building on this, introducing semantic latents further accelerates performance improvement, while Unified Latent consistently outperforms Independent Latents. This indicates that the benefit does not simply come from adding multiple semantic prediction objectives, but from jointly encoding correlated manipulation knowledge across semantic levels into an internal representation for action generation. In particular, Unified Latent reaches 73.58\% at 60K training steps, already surpassing the 72.92\% achieved by the Baseline at 100K steps, attaining a higher success rate with approximately 40\% fewer training steps. These results show that the unified hierarchical semantic representation improves not only final policy performance but also the efficiency with which semantic supervision is translated into action learning.

\subsection{Cross-Embodiment Semantic Retrieval}
\label{sec:cross_embodiment_retrieval}

\begin{wrapfigure}{r}{0.5\linewidth}
    \vspace{-10pt}
    \centering
    \includegraphics[width=\linewidth]{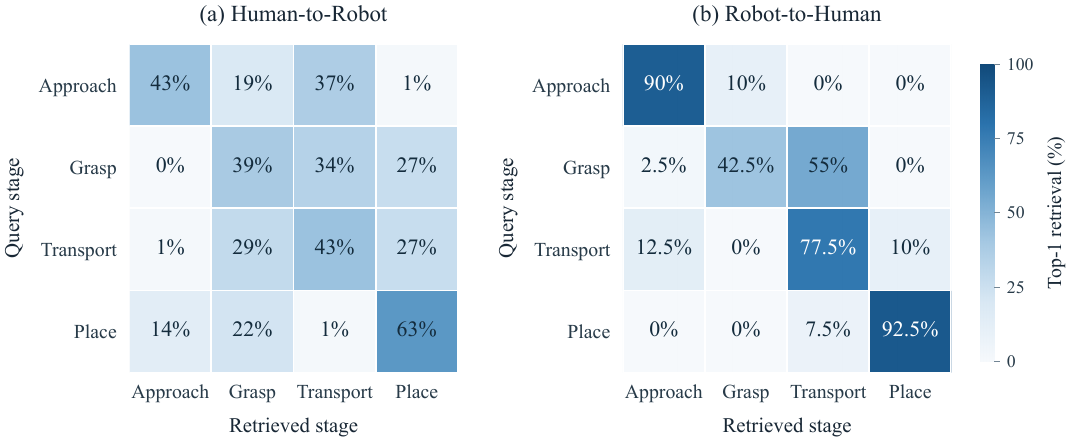}
    \caption{
    Bidirectional cross-embodiment manipulation-stage retrieval using the Hierarchical Semantic Latent. Rows represent query stages and columns represent top-1 retrieved stages, with diagonal entries indicating correct matches.
    }
    \label{fig:cross_embodiment_retrieval}
    \vspace{-10pt}
\end{wrapfigure}

To examine whether the Hierarchical Semantic Latent captures manipulation semantics shared across embodiments, we conduct bidirectional cross-embodiment retrieval. We select human and robot demonstrations from different manipulation tasks that share four common stages: Approach, Grasp, Transport, and Place. Given a query frame from one embodiment, we retrieve semantically similar representations from the other embodiment, treating samples from the same stage as positives. Detailed evaluation protocols are provided in Appendix~\ref{sec:retrieval_details}.

As shown in Figure~\ref{fig:cross_embodiment_retrieval}, the learned representation achieves R@1 scores of 47.00\% for Human-to-Robot retrieval and 75.63\% for Robot-to-Human retrieval, both exceeding the 25\% chance level. In particular, the Place stage achieves 63.0\% and 92.5\% in the two directions, respectively, while Approach and Transport reach 90.0\% and 77.5\% in Robot-to-Human retrieval. These results demonstrate that the Hierarchical Semantic Latent captures shared manipulation-stage semantics across different human and robot tasks, providing representation-level evidence of cross-embodiment semantic correspondence.

\subsection{Human Data Scaling}
\label{sec:human_data_scaling}

\begin{wrapfigure}{r}{0.5\linewidth}
    \centering
    \includegraphics[width=\linewidth]{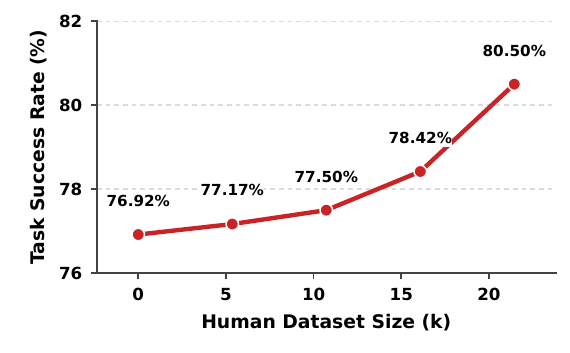}
    \caption{Human data scaling on RoboCasa-GR1 using nested subsets of filtered EgoDex demonstrations. The zero-data point corresponds to the robot-only reference; co-training uses a fixed human-to-robot sampling ratio of 1:2.}
    \label{fig:human_data_scaling}
\end{wrapfigure}

We further investigate how the size of the human demonstration dataset affects robot policy performance. Using the filtered EgoDex~\citep{hoque2026egodex} dataset, we construct nested subsets containing approximately 5.4k, 10.7k, 16.1k, and 21.4k human demonstrations. All human--robot co-training variants use a fixed human-to-robot sampling ratio of 1:2 and are evaluated at 150K training steps. The robot-only reference retains both hierarchical semantic supervision and the unified latent representation but excludes human demonstrations, and is evaluated at 100K steps. With the same global batch size, this protocol matches the expected robot-data exposure across variants. Comparisons among the co-training variants therefore isolate the effect of expanding the human demonstration pool while keeping the training budget and sampling ratio fixed.

As shown in Figure~\ref{fig:human_data_scaling}, the observed success rates increase with the size of the human dataset, reaching 80.50\% with all 21,446 demonstrations, a 3.58 percentage-point improvement over the robot-only reference. Among the co-training variants, increasing the human dataset from approximately 5.4k to 21.4k demonstrations improves the success rate from 77.17\% to 80.50\%, a gain of 3.33 percentage points. The largest observed incremental gain occurs when expanding the dataset from approximately 16.1k to 21.4k demonstrations, improving the success rate by 2.08 percentage points. Since all co-training variants use the same training budget and sampling ratio, these results suggest that access to a larger pool of human demonstrations benefits robot policy learning within the evaluated range.

\subsection{Real-World Experiments}
\label{sec:real_world}

We further evaluate Nexus-VLA on real-world robot manipulation tasks and construct four controlled OOD settings. Object Distractor introduces similar or irrelevant objects to interfere with target selection; Spatial Layout changes the spatial configuration between objects and targets; Motion Constraint introduces obstacles that alter feasible motion paths; and Unseen Object introduces manipulation objects absent from robot training. These settings probe semantic generalization along dimensions related to Intent, Spatial, Motion, and Interaction.

As shown in Table~\ref{tab:real_world}, Nexus-VLA achieves the highest success rate under both the standard setting and all four OOD settings, with an average success rate of 74.0\%. Human-Robot Co-training alone provides only a modest improvement, whereas adding Hierarchical Alignment and Unified Latent leads to more substantial gains under OOD conditions, with the full Nexus-VLA achieving the best overall performance. Compared with the Baseline, the performance gap is 12.0 percentage points under the standard setting and 22.0--26.0 percentage points across the four OOD settings. This larger gap under distribution shifts suggests that Nexus-VLA not only improves standard-task performance but also learns representations that transfer more robustly across changes in targets, spatial layouts, motion constraints, and object instances.

\begin{wrapfigure}{r}{0.55\linewidth}
    \centering
    \includegraphics[width=\linewidth]{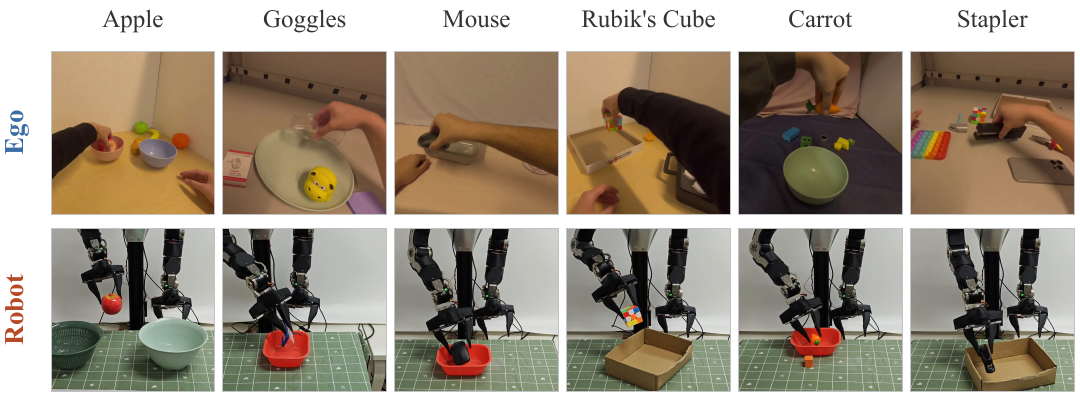}
    \caption{Cross-embodiment transfer from human experience to robot execution.}
    \label{fig:ego2robot}
\end{wrapfigure}

We further investigate whether manipulation experience from human videos can transfer to real-world objects absent from robot training. To this end, we select six object categories from EgoDex that appear only in human demonstrations and construct corresponding real-world robot manipulation tasks, as shown in Figure~\ref{fig:ego2robot}. Each task is evaluated over 50 rollouts. The Robot-only Baseline achieves an average success rate of 47.0\%, whereas Nexus-VLA reaches 72.7\%, a difference of 25.7 percentage points. This result provides direct evidence that human-side manipulation experience can transfer to real-world robot execution even without corresponding robot demonstrations.

\begin{table}[t]
\centering
\caption{
Real-world manipulation and OOD generalization.
We evaluate six real-world manipulation tasks under the standard setting
and four controlled distribution shifts.
Each setting is evaluated over 50 robot rollouts.
All numbers are success rates (\%).
}
\label{tab:real_world}
\resizebox{\linewidth}{!}{
\begin{tabular}{lcccccc}
\toprule
Method
& Standard
& Object Distractor
& Spatial Layout
& Motion Constraint
& Unseen Object
& Avg. \\
\midrule

$\pi_{0.5}$
& 64.0
& 48.0
& 50.0
& 44.0
& 40.0
& 49.2 \\

GR00T-N1.6
& 58.0
& 42.0
& 44.0
& 38.0
& 34.0
& 43.2 \\

\midrule

Baseline
& 70.0
& 50.0
& 52.0
& 46.0
& 42.0
& 52.0 \\

+ Human-Robot Co-training
& 74.0
& 56.0
& 58.0
& 52.0
& 48.0
& 57.6 \\

+ Hierarchical Alignment + Unified Latent
& 78.0
& 68.0
& 66.0
& 62.0
& 58.0
& 66.4 \\

Nexus-VLA
& 82.0
& 76.0
& 74.0
& 70.0
& 68.0
& 74.0 \\

\bottomrule
\end{tabular}
}
\end{table}

\section{Conclusion}
We present Nexus-VLA, an end-to-end framework for joint learning from unpaired human and robot demonstrations. By aligning cross-embodiment manipulation knowledge across semantic levels and encoding it into a shared Hierarchical Semantic Latent for action generation, Nexus-VLA enables human experience to directly improve robot policy learning. Experiments on RoboCasa-GR1 and real robots show improved manipulation performance, learning efficiency, and generalization. These results show that hierarchical semantic latents effectively transfer scalable human manipulation experience to robot control.

\bibliography{preprint_style}
\bibliographystyle{preprint_style}

\appendix
\section{Appendix}
\subsection{Hierarchical Semantic Supervision and Latent Representation Shaping}
\label{sec:semantic_supervision_details}

As described in the main text, the Hierarchical Semantic Latent is jointly
shaped by four types of structured semantic supervision: Intent, Spatial,
Motion, and Interaction.
This section provides further details on the automatic construction of these
supervision signals, the associated quality-control procedure, and the
semantic projectors and training objectives.
All supervision is generated offline by frozen Structured Manipulation Experts
(SMEs) and used to shape the shared Semantic Latent through four lightweight
trainable projectors: the Subtask, Grounding, Motion, and Contact Projectors.

% =========================================================
\subsubsection{Structured Manipulation Experts for Automatic Supervision}
\label{sec:structured_expert}

\begin{figure}[t]
    \centering
    \includegraphics[width=1\linewidth]{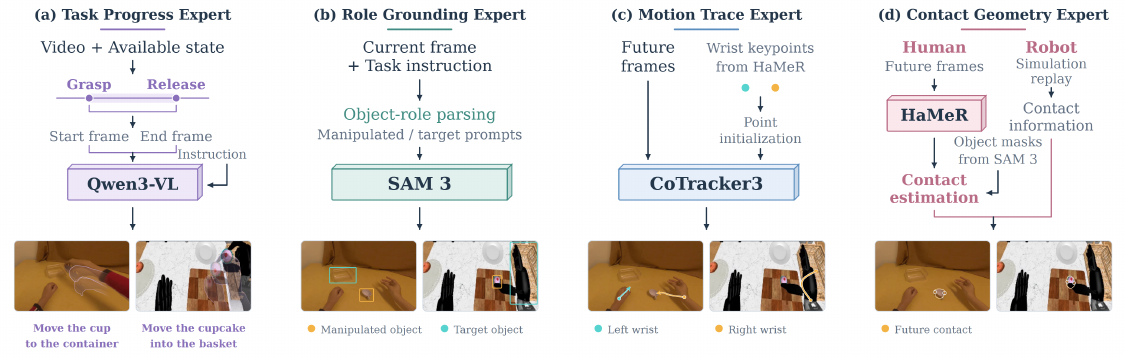}
    \caption{
    Overview of the four Structured Manipulation Experts for
    constructing Intent, Spatial, Motion, and Interaction supervision.
    All SMEs are frozen and used only for offline supervision generation.
    }
    \label{fig:sme_experts_detail}
\end{figure}

At timestep $t$, the structured supervision generated by the SMEs is denoted as
\begin{equation}
\mathcal{Y}_t
=
\left\{
\mathbf y_t^{\mathrm{intent}},
\mathbf y_t^{\mathrm{spatial}},
\mathbf y_t^{\mathrm{motion}},
\mathbf y_t^{\mathrm{interaction}}
\right\},
\label{eq:sme_targets}
\end{equation}
which describes the current manipulation intent, object-level spatial
information, future motion, and local physical interaction, respectively.
Although human and robot demonstrations differ in their observations and
execution forms, the four supervision types are represented in semantically
consistent spaces across embodiments, providing unified constraints for
learning shared cross-embodiment representations.
Figure~\ref{fig:sme_experts_detail} summarizes the four offline SMEs and their
corresponding supervision-construction pipelines.

\medskip
\noindent\textbf{Task Progress Expert (TPE).}
TPE constructs the Intent supervision.
We identify key interaction events from changes in hand or gripper opening
and closing, and feed the start frame, end frame, and task instruction of
each resulting manipulation segment into frozen Qwen3-VL to generate a local
subtask description.
Let the $m$-th manipulation segment be
$\mathcal S_m=[t_m^s,t_m^e]$.
The generated local subtask description is used as the Intent supervision
for all $t\in\mathcal S_m$, such that frames within the same segment share
the same local semantic target.

\medskip
\noindent\textbf{Role Grounding Expert (RGE).}
RGE constructs the Spatial supervision.
Given the task instruction, Qwen3-VL identifies the manipuland and target
involved in the current manipulation and predicts the corresponding visual
anchors.
Frozen SAM~3 expands these anchors into object regions and produces
frame-wise 2D bounding boxes together with visibility annotations.
Specifically,
\begin{equation}
\mathbf y_t^{\mathrm{spatial}}
=
\left\{
\mathbf b_t^{\mathrm{manip}},
v_t^{\mathrm{manip}},
\mathbf b_t^{\mathrm{target}},
v_t^{\mathrm{target}}
\right\},
\label{eq:spatial_supervision}
\end{equation}
where $\mathbf b_t\in\mathbb R^4$ denotes a 2D bounding box and $v_t$
indicates whether the corresponding object is visible in the current view.

\medskip
\noindent\textbf{Motion Trace Expert (MTE).}
MTE constructs the Motion supervision.
For human demonstrations, we obtain the left and right wrist locations from
hand keypoints and use frozen CoTracker3 to track them over the following
$H$ frames:
\begin{equation}
\mathbf y_t^{\mathrm{motion}}
=
\left\{
\mathbf p_{t+\tau}^{L},
\mathbf p_{t+\tau}^{R}
\right\}_{\tau=1}^{H},
\qquad
\mathbf p_{t+\tau}^{h}\in\mathbb R^2,
\label{eq:wrist_trajectory}
\end{equation}
where $h\in\{L,R\}$ denotes the left or right wrist.
For robot demonstrations, the corresponding trajectories are obtained from
the end-effector states and converted into the same supervision representation.

\medskip
\noindent\textbf{Contact Geometry Expert (CGE).}
CGE constructs the Interaction supervision.
For human demonstrations, we project the 3D hand keypoints recovered by HaMeR
into the image plane and combine them with the manipuland mask provided by RGE
to localize hand-object contact locations.
These locations are used as contact anchors and propagated temporally to
obtain future contact points and contact-state supervision for frames
preceding the interaction.
For robot demonstrations, the corresponding supervision is constructed from
contact information obtained through simulation replay and converted into the
same visual representation.
The unified Interaction supervision is written as
\begin{equation}
\mathbf y_t^{\mathrm{interaction}}
=
\left\{
\mathcal C_t^L,s_t^L,
\mathcal C_t^R,s_t^R
\right\},
\label{eq:interaction_supervision}
\end{equation}
where $\mathcal C_t^h$ denotes the future contact locations
expressed in the current image coordinates, and $s_t^h$ denotes the
corresponding contact state.

The SMEs are used only to construct training supervision offline.
During policy inference, none of these experts are executed, nor are the
corresponding intermediate semantics explicitly generated.

% =========================================================
\subsubsection{Data Quality Control and Low-Quality Sample Filtering}
\label{sec:data_filtering}

Because the structured supervision is generated automatically by offline experts, we apply hierarchical quality control and conservative filtering to reduce the effects of detection errors, tracking drift, and temporal noise.
For recoverable local errors, we first rerun object localization and segmentation; a demonstration is discarded only when the corresponding supervision cannot be reliably recovered.

Quality control focuses primarily on visual annotation quality and temporal consistency.
For visual annotations, we examine the completeness and temporal stability of the manipuland, target, and hand masks, including persistent missing regions, abrupt changes in position or scale, and tracking results that are inconsistent with the observed manipulation motion.
Samples that remain ambiguous after these checks are further reviewed using a frozen vision-language model.
For temporal consistency, we verify key interaction events, active-hand tracking, and the temporal alignment among video observations, state trajectories, and structured supervision.

Demonstrations that fail these quality checks and cannot be reliably recovered are removed at the episode level.
For otherwise valid demonstrations with locally missing supervision, the corresponding invalid targets are masked out using validity masks while the remaining reliable supervision is retained.

% =========================================================
\subsubsection{Semantic projectors and Training Objectives}
\label{sec:semantic_heads_details}

The Hierarchical Semantic Latent does not have explicit latent-level ground truth.
Instead, we decode four types of structured semantics from the same
$\mathbf z_t^{\mathrm{sem}}$ and jointly shape its representation using the corresponding prediction losses.
To encourage the semantic information to be encoded primarily in the shared latent representation,
all four prediction modules use lightweight architectures, as summarized in Table~\ref{tab:semantic_heads}.

\begin{table}[t]
\centering
\caption{Architecture of the hierarchical semantic projectors.}
\label{tab:semantic_heads}
\small
\setlength{\tabcolsep}{4pt}
\renewcommand{\arraystretch}{1.12}

\begin{tabularx}{\linewidth}{
    @{}l
    >{\raggedright\arraybackslash}X
    >{\raggedright\arraybackslash}X@{}
}
\toprule
\textbf{Projector}
& \textbf{Architecture}
& \textbf{Semantic Target} \\
\midrule

Subtask Projector
& Linear ($2048{\rightarrow}768$) + 1-layer causal Transformer
& Local subtask \\

Grounding Projector
& MLP ($2048{\rightarrow}512{\rightarrow}8$) + visibility head
& Manipuland-target grounding \\

Motion Projector
& MLP ($2048{\rightarrow}1024{\rightarrow}64$) + visibility head
& Future bimanual wrist trajectory \\

Contact Projector
& MLP ($2048{\rightarrow}512{\rightarrow}8$) + existence head
& Hand-object contact geometry \\

\bottomrule
\end{tabularx}
\end{table}

To handle missing annotations, invisible targets, and invalid timesteps in a unified manner,
we define the masked average as
\begin{equation}
\left\langle \ell_i \right\rangle_{\mathbf m}
=
\frac{\sum_i m_i\,\ell_i}
{\sum_i m_i+\epsilon},
\label{eq:masked_average}
\end{equation}
where $\mathbf m$ denotes the validity mask associated with the corresponding supervision.

\medskip
\noindent\textbf{Intent.}
The Subtask Projector autoregressively predicts the local subtask description conditioned on
$\mathbf z_t^{\mathrm{sem}}$.
Given the target token sequence
$\mathbf y_t^{\mathrm{intent}}=(y_1,\ldots,y_M)$,
the training objective is
\begin{equation}
\mathcal L_{\mathrm{intent}}
=
\left\langle
-\log
p\!\left(
y_j\mid y_{<j},\mathbf z_t^{\mathrm{sem}}
\right)
\right\rangle_{\mathbf m^{\mathrm{tok}}},
\label{eq:intent_loss}
\end{equation}
where $\mathbf m^{\mathrm{tok}}$ masks padding tokens and samples without valid text supervision.

\medskip
\noindent\textbf{Spatial.}
The Grounding Projector predicts the 2D bounding boxes of the manipuland and target together with their visibility.
Let $\mathbf a$ and $\mathbf v$ denote annotation validity and visibility, respectively. Let $r\in\{\mathrm{manip},\mathrm{target}\}$ index the object role.
The corresponding loss is
\begin{align}
\mathcal L_{\mathrm{spatial}}
={}&
\left\langle
\mathcal L_{\mathrm{box}}
\!\left(
\hat{\mathbf b}_t^r,\mathbf b_t^r
\right)
\right\rangle_{\mathbf a\odot\mathbf v}
\nonumber\\
&+
\left\langle
\operatorname{BCE}
\!\left(
\hat v_t^r,v_t^r
\right)
\right\rangle_{\mathbf a},
\label{eq:spatial_loss}
\end{align}
where
\begin{equation}
\mathcal L_{\mathrm{box}}
=
\operatorname{SmoothL1}
+
\mathcal L_{\mathrm{GIoU}}.
\label{eq:bbox_regression_loss}
\end{equation}

\medskip
\noindent\textbf{Motion.}
The Motion Projector predicts the future 2D trajectories of both wrists over
$H=16$ frames, together with their visibility:
\begin{align}
\mathcal L_{\mathrm{motion}}
={}&
\left\langle
\operatorname{SmoothL1}
\!\left(
\hat{\mathbf p}_{t+\tau}^{h},
\mathbf p_{t+\tau}^{h}
\right)
\right\rangle_{\mathbf m^{\mathrm{traj}}\odot\mathbf v}
\nonumber\\
&+
\left\langle
\operatorname{BCE}
\!\left(
\hat v_{t+\tau}^{h},
v_{t+\tau}^{h}
\right)
\right\rangle_{\mathbf m^{\mathrm{traj}}},
\label{eq:motion_loss}
\end{align}
where $h\in\{L,R\}$, $\tau=1,\ldots,H$, and
$\mathbf m^{\mathrm{traj}}$ denotes the validity mask for trajectory supervision.

\medskip
\noindent\textbf{Interaction.}
The Contact Projector predicts key contact locations for both hands together with the corresponding contact-existence states:
\begin{align}
\mathcal L_{\mathrm{interaction}}
={}&
\left\langle
\operatorname{SmoothL1}
\!\left(
\hat{\mathbf c}_t^{h,q},
\mathbf c_t^{h,q}
\right)
\right\rangle_{\mathbf p}
\nonumber\\
&+
\left\langle
\operatorname{BCE}
\!\left(
\hat s_t^h,s_t^h
\right)
\right\rangle_{\mathbf a},
\label{eq:interaction_loss}
\end{align}
where $q$ indexes contact roles within each hand,
and $\mathbf p$ and $\mathbf a$ denote the validity masks for contact points and Interaction annotations, respectively.

The four prediction losses jointly define the semantic shaping objective:
\begin{equation}
\mathcal L_{\mathrm{sem}}
=
\sum_{k\in\mathcal K}
\lambda_k \mathcal L_k,
\qquad
\mathcal K=
\{
\mathrm{intent},
\mathrm{spatial},
\mathrm{motion},
\mathrm{interaction}
\}.
\label{eq:semantic_loss_detail}
\end{equation}
During training, we set all $\lambda_k=1.0$ and
$\lambda_{\mathrm{sem}}=0.1$, and jointly optimize the semantic shaping
objective with the action-generation objective:
\begin{equation}
\mathcal L_{\mathrm{joint}}
=
\mathcal L_{\mathrm{FM}}
+
\lambda_{\mathrm{sem}}
\mathcal L_{\mathrm{sem}}.
\label{eq:joint_loss_detail}
\end{equation}

These semantic projectors are used only to shape the representation during training.
At inference time, the policy directly uses the semantically shaped
$\mathbf z_t^{\mathrm{sem}}$ for action generation without explicitly decoding any of the intermediate semantics above.

\begin{table*}[t]
\centering
\caption{Training and optimization configuration of Nexus-VLA.}
\label{tab:training_hyperparameters}
\small
\setlength{\tabcolsep}{7pt}
\renewcommand{\arraystretch}{1.10}

\begin{tabularx}{\textwidth}{
    @{}l >{\centering\arraybackslash}X
    l >{\centering\arraybackslash}X@{}
}
\toprule
\textbf{Configuration} & \textbf{Value}
& \textbf{Configuration} & \textbf{Value} \\
\midrule

Action horizon
& 16
& Training steps
& 200K \\

Per-GPU batch size
& 32
& Gradient accumulation
& 1 \\

Global batch size
& 512
& Learning rate
& $3\times10^{-5}$ \\

Optimizer
& AdamW
& LR schedule
& Cosine decay \\

Warmup ratio
& 0.05
& Weight decay
& $1\times10^{-5}$ \\

Human : Robot sampling
& $1:2$
& Semantic branch weights $\lambda_k$
& $1.0$ \\

Semantic weight $\lambda_{\mathrm{sem}}$
& 0.1
& Training precision
& BF16 \\

\bottomrule
\end{tabularx}
\end{table*}

\begin{table*}[t]
\centering
\caption{Architecture configuration of Nexus-VLA.}
\label{tab:model_architecture}
\small
\setlength{\tabcolsep}{8pt}
\renewcommand{\arraystretch}{1.08}

\begin{tabularx}{\textwidth}{
    @{}l
    >{\centering\arraybackslash}X
    >{\centering\arraybackslash}X@{}
}
\toprule
\textbf{Configuration}
& \textbf{VLM Backbone}
& \textbf{Action Head} \\
\midrule

Backbone
& Qwen2.5-VL-3B-Instruct
& Flow-Matching DiT \\

Transformer layers
& 36 (LLM) / 32 (ViT)
& 32 \\

Residual width
& 2048 / 1280
& 1536 \\

Attention heads $\times$ dim.
& $16\times128$ / $16\times80$
& $32\times48$ \\

Positional encoding
& M-RoPE
& Learned 1D + sinusoidal time \\

FFN intermediate width
& 11008 / 3420
& 6144 \\

Input
& Image + text + $\mathbf q_{\mathrm{sem}}$
& $\mathbf s_t$, $\mathbf A_t^\tau$, $\tau$, $\mathbf H_t$ \\

Output
& $\mathbf H_t$, $\mathbf z_t^{\mathrm{sem}}$
& $\hat{\mathbf v}_t^\tau$ \\

\bottomrule
\end{tabularx}
\end{table*}

\section{Model and Training Implementation Details}
\label{sec:implementation_details}

This section provides additional details on the Human-Robot joint training strategy and the main optimization settings.

% =========================================================
\subsection{Human-Robot Joint Training}
\label{sec:human_robot_training}

To jointly leverage human and robot demonstrations within a single policy, we adopt a unified action representation and write the robot action as the concatenation of joint-space and wrist-space components:
\begin{equation}
\mathbf a_t
=
\left[
\mathbf a_t^{\mathrm{joint}};
\mathbf a_t^{\mathrm{wrist}}
\right].
\label{eq:unified_action}
\end{equation}
Robot demonstrations contain a 29-dimensional joint action and an 18-dimensional bimanual wrist pose, resulting in a 47-dimensional action representation.
For human demonstrations, we use frozen HaMeR to recover the left- and right-wrist poses, each represented by a 3D position and a 6D rotation, yielding the shared 18-dimensional wrist-space action.
Only these shared dimensions participate in action learning, while the robot-specific joint-space dimensions are masked out using an action validity mask.
This allows human and robot data to share the same action-generation module without requiring nonexistent robot joint actions to be constructed for human demonstrations.

To further control the contribution of different embodiments during joint training, we adopt hierarchical probabilistic sampling.
For a data source
$d\in\{\mathrm{human},\mathrm{robot}\}$,
let $\mathcal D_d$ denote the set of datasets belonging to that source, and let $n_j$ denote the number of samples in dataset $j$.
The sampling weight of dataset $j$ is defined as
\begin{equation}
w_j
=
s_d
\frac{n_j}
{\sum_{k\in\mathcal D_d}n_k},
\qquad
j\in\mathcal D_d,
\qquad
\sum_d s_d=1.
\label{eq:dataset_sampling}
\end{equation}
Here, $s_d$ controls the overall sampling proportion of each embodiment, while datasets within the same source are sampled in proportion to their sizes.
In our experiments, the Human : Robot sampling ratio is set to $1:2$, preventing the relative contribution of the two data sources from being determined directly by their original dataset sizes.

\begin{table*}[t]
\centering
\caption{Per-task success rates (\%) on all 24 RoboCasa-GR1 tasks. Tasks
follow the Close, Cuttingboard, Placemat, Tray, and Plate group ordering in
Table~\ref{tab:robocasa_main}. Best and second-best values are boldfaced
and underlined, respectively; ties receive the same formatting.}
\label{tab:robocasa_per_task}
\footnotesize
\setlength{\tabcolsep}{6pt}
\renewcommand{\arraystretch}{1.06}
\begin{tabularx}{\textwidth}{
    @{}
    >{\raggedright\arraybackslash}p{0.36\textwidth}
    *{4}{>{\centering\arraybackslash}X}
    @{}
}
\toprule
\textbf{Task}
& \textbf{ZR-0}
& \mbox{\textbf{JoyAI-RA 0.1}}
& \textbf{ACE-Ego-0}
& \textbf{Nexus-VLA} \\
\midrule
CupToDrawerClose & 20.0 & \textbf{48.0} & \underline{36.0} & \textbf{48.0} \\
PotatoToMicrowaveClose & 59.0 & \underline{70.0} & 58.0 & \textbf{80.0} \\
MilkToMicrowaveClose & 45.0 & \textbf{84.0} & 56.0 & \underline{74.0} \\
BottleToCabinetClose & 39.0 & \underline{84.0} & 78.0 & \textbf{88.0} \\
WineToCabinetClose & 40.0 & \underline{54.0} & \textbf{56.0} & \textbf{56.0} \\
CanToDrawerClose & 47.0 & \textbf{90.0} & 70.0 & \underline{76.0} \\
\midrule
CuttingboardToBasket & 82.0 & \underline{88.0} & 84.0 & \textbf{92.0} \\
CuttingboardToCardboardBox & 79.0 & 46.0 & \underline{84.0} & \textbf{94.0} \\
CuttingboardToPan & \underline{92.0} & \underline{92.0} & \underline{92.0} & \textbf{96.0} \\
CuttingboardToPot & \underline{85.0} & 80.0 & 84.0 & \textbf{88.0} \\
CuttingboardToTieredBasket & \textbf{80.0} & 36.0 & 72.0 & \underline{78.0} \\
\midrule
PlacematToBasket & 78.0 & 76.0 & \underline{86.0} & \textbf{90.0} \\
PlacematToBowl & \underline{87.0} & 52.0 & 72.0 & \textbf{92.0} \\
PlacematToPlate & \underline{88.0} & 38.0 & 80.0 & \textbf{90.0} \\
PlacematToTieredShelf & \underline{46.0} & 14.0 & 44.0 & \textbf{48.0} \\
\midrule
TrayToCardboardBox & 81.0 & \underline{82.0} & 78.0 & \textbf{86.0} \\
TrayToPlate & 81.0 & 88.0 & \underline{90.0} & \textbf{92.0} \\
TrayToPot & 74.0 & 88.0 & \textbf{98.0} & \underline{90.0} \\
TrayToTieredBasket & \underline{74.0} & 62.0 & \underline{74.0} & \textbf{76.0} \\
TrayToTieredShelf & \underline{50.0} & 24.0 & \underline{50.0} & \textbf{74.0} \\
\midrule
PlateToBowl & \underline{82.0} & 48.0 & 68.0 & \textbf{88.0} \\
PlateToCardboardBox & \underline{81.0} & 38.0 & 70.0 & \textbf{84.0} \\
PlateToPan & \textbf{89.0} & 46.0 & \underline{70.0} & \underline{70.0} \\
PlateToPlate & 85.0 & \underline{88.0} & \textbf{98.0} & 82.0 \\
\midrule
\textbf{Average} & 69.3 & 63.2 & \underline{72.8} & \textbf{80.5} \\
\bottomrule
\end{tabularx}
\end{table*}

% =========================================================
\subsection{Model Architecture and Training Configuration}
\label{sec:training_details}

For reproducibility, Table~\ref{tab:model_architecture} summarizes the main architecture of Nexus-VLA, while Table~\ref{tab:training_hyperparameters} reports the training and optimization settings.
Unless otherwise specified, all experiments use the same model architecture and training configuration.
The architectures and supervision targets of the semantic projectors are detailed in Section~\ref{sec:semantic_heads_details}.

\begin{figure}[t]
    \centering
        \includegraphics[width=1\linewidth]{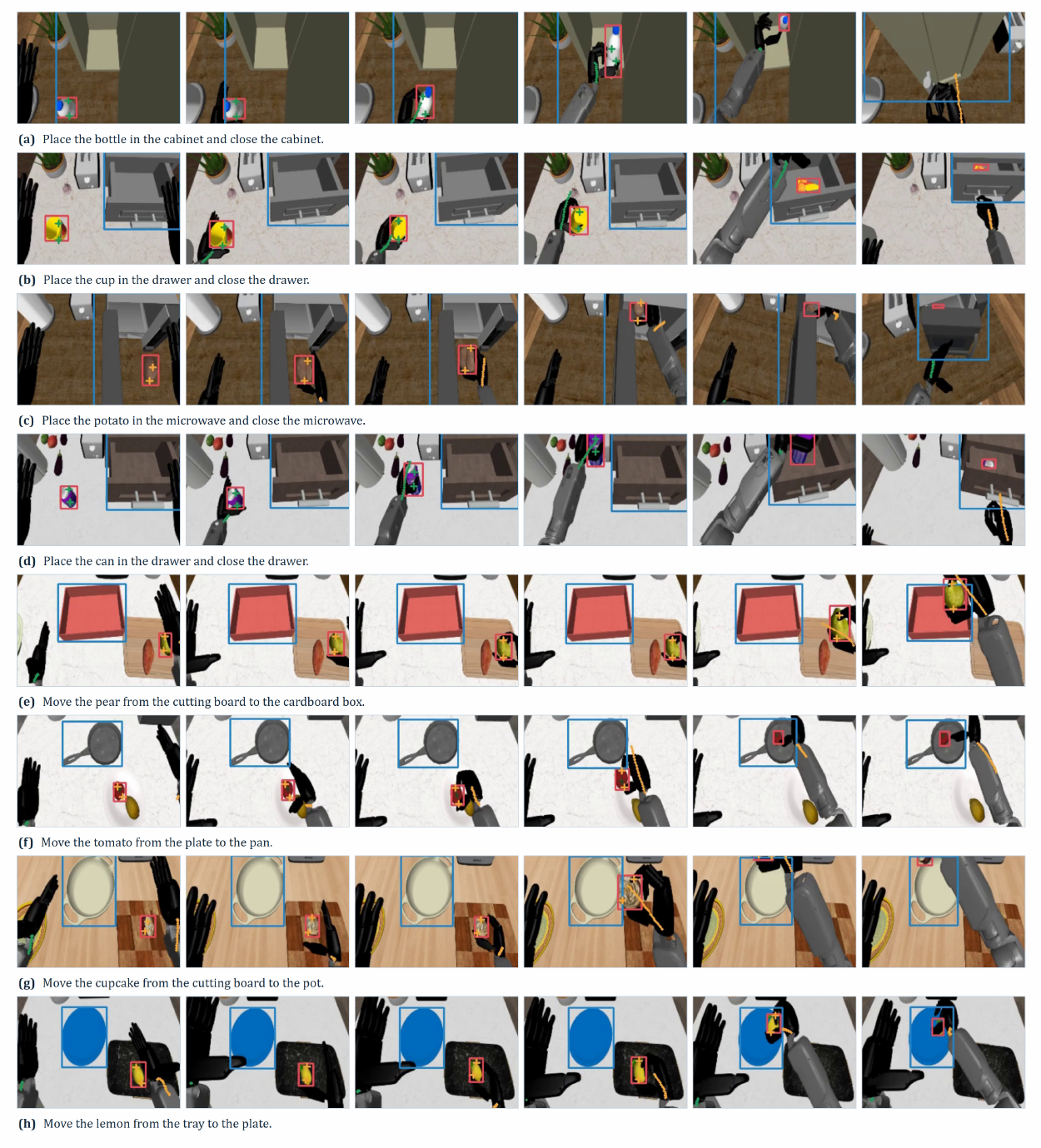}
        
\caption{Qualitative visualization of eight RoboCasa-GR1 manipulation tasks. Each row shows six time-ordered frames, with colored overlays highlighting task-relevant objects, targets, and motion or interaction cues.}
    \label{fig:supp_robocasa_latent}
\end{figure}

\section{Supplementary Experimental Results}
\label{sec:supp_results}

\subsection{Additional RoboCasa-GR1 Results}
\label{sec:supp_robocasa}

RoboCasa-GR1 is our primary simulation benchmark and contains 24 humanoid
manipulation tasks executed by the Fourier GR-1 robot. The task suite covers
both object-transfer behaviors and interactions with articulated structures,
including drawers, cabinets, and microwaves. In addition to the standard test
distribution, we construct six controlled OOD settings to evaluate
generalization under changes in visual appearance, spatial configuration,
scene composition, and object identity.

Specifically, \textbf{Illumination} and \textbf{Visual Noise} perturb the
visual observations while preserving the underlying task structure.
\textbf{Cross-Configuration} evaluates novel spatial arrangements of
task-relevant entities, while \textbf{Object-Receptacle} introduces
object-target compositions not observed in the standard setting.
We further consider two object-level shifts:
\textbf{Ego-Object} uses objects absent from robot training but observed in
the human demonstrations, whereas \textbf{Fully Unseen} uses objects that
appear in neither the human nor robot training data.
These settings complement the standard benchmark by progressively testing
robustness to perceptual, spatial, compositional, and object-level
distribution shifts.

Table~\ref{tab:robocasa_per_task} expands the five task-group results in
Table~\ref{tab:robocasa_main} into success rates for all 24 RoboCasa-GR1
tasks. Nexus-VLA achieves the highest rate, including ties, on 18 tasks and
the second-highest rate on five more. Its average success rate is 80.5\%,
7.7 percentage points above ACE-Ego-0, the strongest competing method by
overall average. The per-task results show that this improvement spans
articulated-object tasks and several object-transfer groups rather than
being driven by a single task family. The gains are especially clear on PlacematToBowl (92.0\% versus 72.0\% for
ACE-Ego-0) and TrayToTieredShelf (74.0\% versus 50.0\%). Nexus-VLA also
attains the best result on four of the six Close tasks, including ties,
although JoyAI-RA 0.1 has a slightly higher Close-group average (71.7\%
versus 70.3\%). The advantage is therefore not uniform: ZR-0 performs
better on PlateToPan, and ACE-Ego-0 performs better on PlateToPlate. These
cases identify task-level variation that is obscured by the aggregate
benchmark score.

Because the Hierarchical Semantic Latent is an internal representation,
Figure~\ref{fig:supp_robocasa_latent} visualizes task-relevant semantic cues
alongside the executed motion rather than depicting the latent as an
explicit symbol sequence. The eight representative RoboCasa-GR1 rollouts
cover both articulated interactions and object transfers. Across six
time-ordered frames per task, the marked object and destination remain
associated with the approach, grasp, transport, placement, and closing
stages as applicable. This qualitative view is consistent with the
coarse-to-fine semantic structure used to shape the latent. The visual
overlays serve interpretation; action generation does not require explicit
intermediate semantic outputs at inference time, as described in
Section~\ref{sec:latent_action_generation}.

\begin{figure}[t]
    \centering
        \includegraphics[width=1\linewidth]{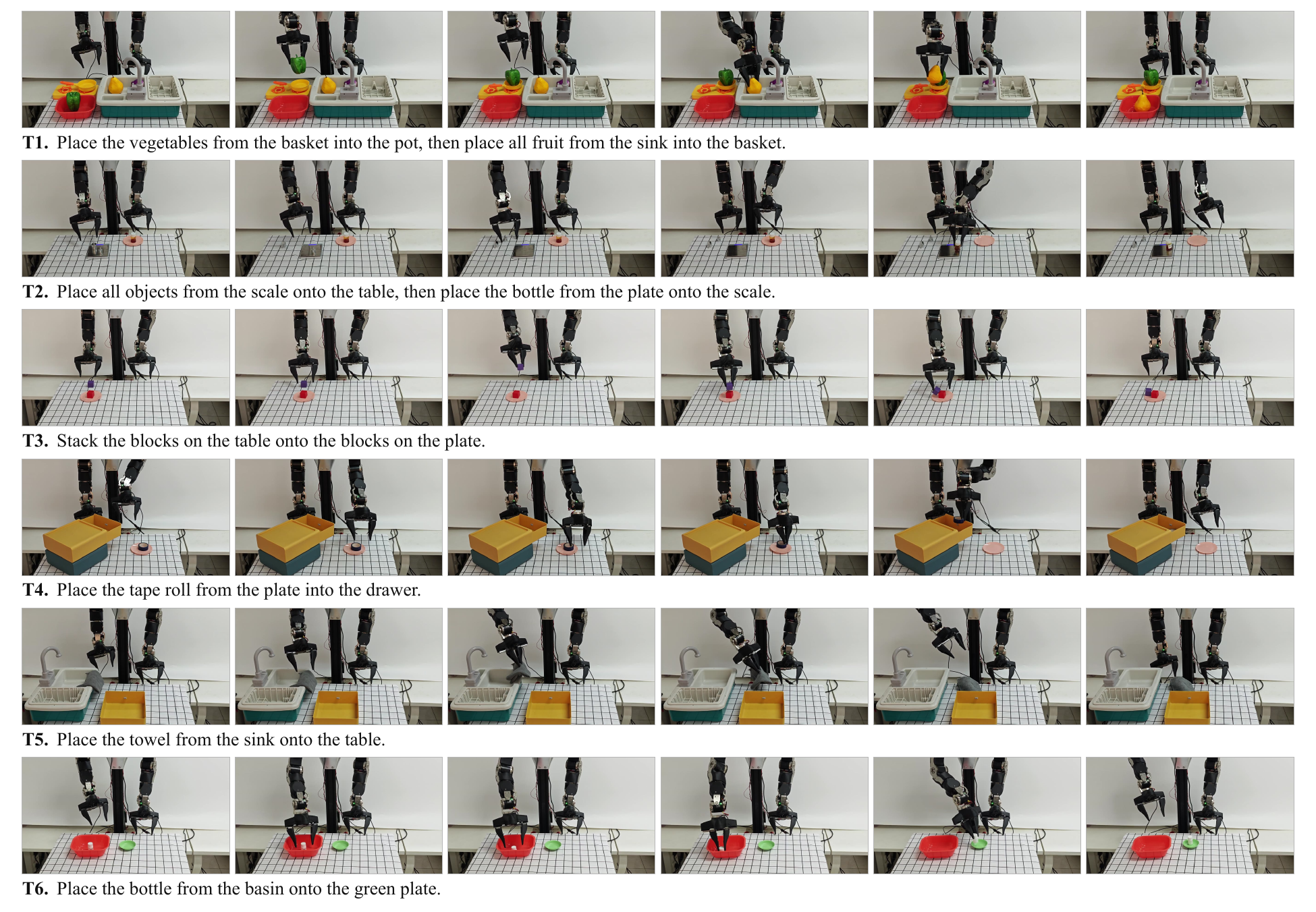}
    \caption{Real-robot executions under the standard setting. Rows T1--T6
    correspond to the six task instructions printed below the image
    sequences. }
    \label{fig:supp_real_standard}
\end{figure}

\subsection{Cross-Embodiment Semantic Retrieval Details}
\label{sec:retrieval_details}

\paragraph{Evaluation Data and Stage Sampling.}
We construct the retrieval evaluation using human demonstrations from EgoDex and robot demonstrations from RoboCasa-GR1. Specifically, we select 20 human demonstrations from the \texttt{basic\_pick\_place} task and eight successful robot demonstrations from the \texttt{PnPCupToDrawerClose} task. The human task involves picking up a cup and placing it into a bowl, while the robot task involves placing a cup into a drawer and subsequently closing the drawer. Both tasks involve left-hand manipulation and share four common stages: Approach, Grasp, Transport, and Place.

For each demonstration, we sample five frames from each stage, yielding 20 frames per demonstration. The resulting evaluation set contains 400 human frames and 160 robot frames, with 100 and 40 samples per stage, respectively. The drawer-closing stage is excluded from the robot demonstrations to maintain a consistent four-stage evaluation protocol.

\paragraph{Semantic Representation Extraction.}
We use the Nexus-VLA checkpoint at 150K training steps to extract the Hierarchical Semantic Latent. Each sampled frame is independently encoded using its visual observation and original task instruction. The 2048-dimensional hidden state corresponding to the semantic token is used as the retrieval representation:
\begin{equation}
\mathbf z_i^{(e)}
=
F_{\theta}^{\mathrm{sem}}
\left(
\mathbf o_i^{(e)},
\ell_i^{(e)}
\right)
\in \mathbb R^{2048},
\end{equation}
where $e\in\{\mathrm{human},\mathrm{robot}\}$ denotes the embodiment, $\mathbf o_i^{(e)}$ represents the visual observation, and $\ell_i^{(e)}$ is the original task instruction. Stage labels are used exclusively for evaluating retrieval relevance and are not provided as encoding inputs.

\paragraph{Feature Processing and Retrieval Protocol.}
For cross-embodiment retrieval, we first apply L2 normalization to each semantic representation:
\begin{equation}
\hat{\mathbf z}_i^{(e)}
=
\frac{\mathbf z_i^{(e)}}
{\|\mathbf z_i^{(e)}\|_2}.
\end{equation}

We then calculate the mean representation separately for each embodiment:
\begin{equation}
\boldsymbol{\mu}_e
=
\frac{1}{N_e}
\sum_{i=1}^{N_e}
\hat{\mathbf z}_i^{(e)},
\end{equation}
where $N_e$ denotes the number of evaluation samples from embodiment $e$. The domain means are estimated from the complete evaluation representations without using stage labels.

Each representation is subsequently centered by its corresponding domain mean and normalized again:
\begin{equation}
\widetilde{\mathbf z}_i^{(e)}
=
\frac{
\hat{\mathbf z}_i^{(e)}-\boldsymbol{\mu}_e
}{
\left\|
\hat{\mathbf z}_i^{(e)}-\boldsymbol{\mu}_e
\right\|_2
}.
\end{equation}

Cross-embodiment retrieval ranks candidates according to cosine similarity:
\begin{equation}
s(i,j)
=
\left(
\widetilde{\mathbf z}_i^{(e_q)}
\right)^\top
\widetilde{\mathbf z}_j^{(e_c)},
\qquad e_q\neq e_c,
\end{equation}
where $e_q$ and $e_c$ denote the query and candidate embodiments.

For intra-embodiment retrieval, we use the original L2-normalized representations $\hat{\mathbf z}_i^{(e)}$ and exclude all samples belonging to the same demonstration as the query. This evaluates stage-level retrieval across different demonstrations within each embodiment.

\paragraph{Evaluation Metrics and Results.}
We report Recall at 1 (R@1), Precision at 5 (P@5), and mean Average Precision (mAP). A candidate is considered relevant if it belongs to the same manipulation stage as the query. R@1 measures the correctness of the top-ranked candidate, while P@5 measures the proportion of relevant candidates among the top five retrieved samples.

For each query $i$, Average Precision is defined as
\begin{equation}
\mathrm{AP}(i)
=
\frac{1}{N_i^{+}}
\sum_{k=1}^{|\mathcal C_i|}
\mathrm{P@}k(i)\,
\mathrm{rel}_i(k),
\end{equation}
where $\mathcal C_i$ is the candidate set, $N_i^{+}$ denotes the number of relevant candidates, and $\mathrm{rel}_i(k)$ indicates whether the candidate at rank $k$ belongs to the query stage.

Metrics are first averaged over queries within each stage and then equally averaged across the four stages. Since candidate samples are uniformly distributed across stages, the chance-level expectation for R@1 and P@5 is 25\%.

\begin{table}[t]
\centering
\caption{
Intra- and cross-embodiment manipulation-stage retrieval results.
All metrics are reported in percentages (\%).
Intra-embodiment retrieval excludes samples from the query demonstration; cross-embodiment retrieval uses domain-wise mean-centered representations.
}
\label{tab:cross_embodiment_retrieval}
\small
\setlength{\tabcolsep}{6pt}
\renewcommand{\arraystretch}{1.12}
\begin{tabular}{lccccc}
\toprule
\textbf{Retrieval}
& \textbf{Queries}
& \textbf{Candidates}
& \textbf{R@1}
& \textbf{P@5}
& \textbf{mAP} \\
\midrule
Human $\rightarrow$ Human
& 400 & 380 & 88.25 & 83.30 & 51.92 \\
Robot $\rightarrow$ Robot
& 160 & 140 & 90.00 & 84.75 & 79.06 \\
\midrule
Human $\rightarrow$ Robot
& 400 & 160 & 47.00 & 45.15 & 49.07 \\
Robot $\rightarrow$ Human
& 160 & 400 & 75.63 & 70.50 & 49.19 \\
\bottomrule
\end{tabular}
\end{table}

As shown in Table~\ref{tab:cross_embodiment_retrieval}, the Hierarchical Semantic Latent achieves R@1 scores of 88.25\% and 90.00\% for Human-to-Human and Robot-to-Robot retrieval, respectively, demonstrating strong stage discriminability within each embodiment. Cross-embodiment retrieval also achieves above-chance performance in both directions, with P@5 scores of 45.15\% and 70.50\%, respectively.

The stage-wise distributions in Figure~\ref{fig:cross_embodiment_retrieval} further characterize the correspondence between manipulation stages across embodiments. The diagonal entries represent correct stage matches, and their four-stage average equals the corresponding R@1 score. The strongest bidirectional correspondence occurs at the Place stage, while Grasp shows greater overlap with Transport, reflecting the closely related manipulation dynamics of these two stages.

\subsection{Additional Real-World Results}
\label{sec:supp_real}

\begin{figure}[htbp]
    \centering
    \includegraphics[width=1\linewidth]{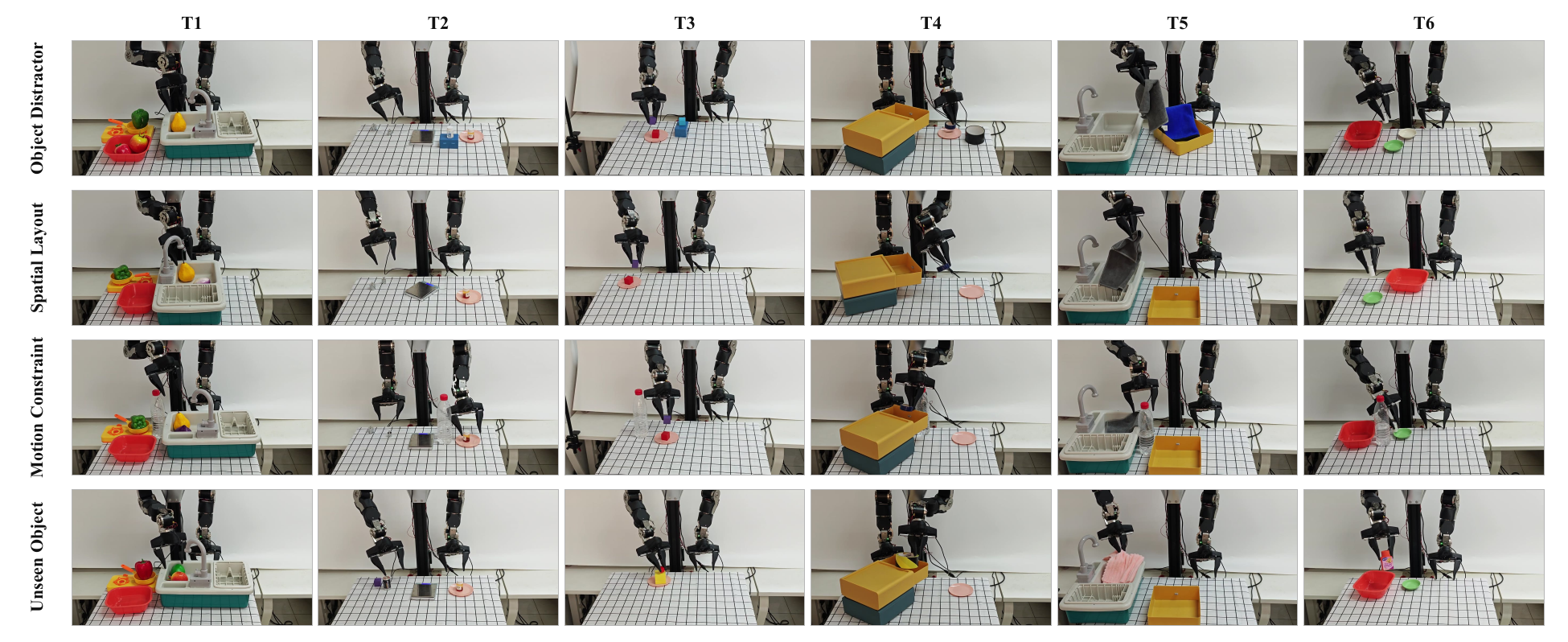}
    \caption{Representative real-robot scenes under four OOD conditions.
    Rows indicate the distribution shift, and columns T1--T6 correspond to
    the standard tasks in Figure~\ref{fig:supp_real_standard}. One frame is
    shown for each condition-task combination.}
    \label{fig:supp_real_ood}
\end{figure}

Figure~\ref{fig:supp_real_standard} presents representative real-robot
executions for the six standard tasks. Each row shows six frames in temporal
order, with the full instruction below the sequence. The examples include
multi-stage transfers, block stacking, placement into a drawer, and moving
objects between a sink or basin and a tabletop target. The frames make the
execution progression visible beyond the success rates reported in
Table~\ref{tab:real_world}.

To further evaluate robustness under controlled distribution shifts, we
construct four real-world OOD settings while preserving the underlying task
objective and success criterion. Each setting primarily stresses a different
aspect of the hierarchical manipulation semantics.

\textbf{Object Distractor} introduces additional objects that are visually
or semantically confusable with task-relevant objects, requiring the policy
to maintain the correct object roles under clutter. For example, we place an
additional tape roll on the table when the instruction specifies the tape on
the plate, introduce plates with distractor colors when the target is the
green plate, or add another towel outside the sink. The intended manipuland
and target remain unchanged.

\textbf{Spatial Layout} changes the relative positions of the manipuland,
target, and surrounding objects while preserving their identities and the
task instruction. This setting evaluates whether the policy can re-ground
task-relevant entities and execute the same manipulation under novel spatial
configurations.

\textbf{Motion Constraint} introduces obstacles along the nominal motion
path without changing the task goal or task-relevant objects. The robot must
therefore adapt its wrist trajectory and use an alternative feasible motion
path to complete the same manipulation.

\textbf{Unseen Object} replaces the manipulated object with an object absent
from the robot training demonstrations while preserving the task structure.
This setting evaluates whether manipulation behavior transfers to novel
object appearances and geometries rather than relying on memorized object
instances.

Following the task ordering in Figure~\ref{fig:supp_real_standard},
Figure~\ref{fig:supp_real_ood} shows representative examples of the four
OOD settings across the same six real-world tasks. Columns T1--T6 correspond
to the six standard tasks, while rows represent Object Distractor, Spatial
Layout, Motion Constraint, and Unseen Object.

\end{document}